\documentclass[sigconf]{acmart}

\usepackage{booktabs} 
\usepackage[caption=false]{subfig}
\usepackage{{adjustbox}}

\renewcommand\footnotetextcopyrightpermission[1]{} 
\acmConference[arXiv version]{}{2018}{July}

\setcopyright{none}

\usepackage[english]{babel}
\usepackage[utf8x]{inputenc}
\usepackage[T1]{fontenc}

\usepackage{amsmath}
\usepackage{multirow}

\newcommand{\alert}[1]{\textcolor{red}{#1}}

\title{Multimodal Social Media Analysis for Gang Violence Prevention}
\author{Philipp Blandfort,\textsuperscript{1,2,}\footnotemark[1]
Desmond Patton,\textsuperscript{3}
William R. Frey,\textsuperscript{3}
Svebor Karaman,\textsuperscript{3}
Surabhi Bhargava,\textsuperscript{3}
Fei-Tzin Lee,\textsuperscript{3}
Siddharth Varia,\textsuperscript{3}
Chris Kedzie,\textsuperscript{3}
Michael B. Gaskell,\textsuperscript{3}
Rossano Schifanella,\textsuperscript{4}
Kathleen McKeown,\textsuperscript{3}
Shih-Fu Chang\textsuperscript{3}}

\affiliation{%
  \institution{\textsuperscript{1} DFKI, Kaiserslautern, Germany\\
  \textsuperscript{2} TU Kaiserslautern, Kaiserslautern, Germany\\
  \textsuperscript{3} Columbia University, New York City, USA\\
  \textsuperscript{4} University of Turin, Turin, Italy}
  }
  
\email{philipp.blandfort@dfki.de, {dp2787, w.frey, svebor.karaman, sb4019, fl2301, sv2504}@columbia.edu, kedzie@cs.columbia.edu, mbg2174@columbia.edu, schifane@di.unito.it, kathy@cs.columbia.edu, sc250@columbia.edu}

\renewcommand{\shortauthors}{P. Blandfort et al.}

\newcommand\blfootnote[1]{%
  \begingroup
  \renewcommand\thefootnote{}\footnote{#1}%
  \addtocounter{footnote}{-1}%
  \endgroup
}

\begin{abstract}
Gang violence is a severe issue in major cities across the U.S. and
recent studies~\cite{patton2017gang} have found evidence of social media communications that can be linked to such violence in communities with high rates of exposure to gang activity. 
In this paper we partnered computer scientists with social work researchers, who have domain expertise in gang violence, to analyze how public tweets with images posted by youth who mention gang associations on Twitter can be leveraged to automatically detect psychosocial factors and conditions that could potentially assist social workers and violence outreach workers in prevention and early intervention programs.
To this end, we developed a rigorous methodology for collecting and annotating tweets.
We gathered 1,851 tweets and accompanying annotations related to visual concepts and the \textit{psychosocial codes}: \textit{aggression}, \textit{loss}, and \textit{substance use}. These codes are relevant to social work interventions, as they represent possible pathways to violence on social media.
We compare various methods for classifying tweets into these three classes, using only the text of the tweet, only the image of the tweet, or both modalities as input to the classifier.
In particular, we analyze the usefulness of mid-level visual concepts and the role of  different modalities for this tweet classification task.
Our experiments show that individually, text information dominates classification performance of the \textit{loss} class, while image information dominates the \textit{aggression} and \textit{substance use} classes. 
Our multimodal approach provides a very promising improvement (18\% relative in mean average precision) over the best single modality approach.
Finally, we also illustrate the complexity of understanding social media data and elaborate on open challenges.
\blfootnote{* During some of this work Blandfort was staying at Columbia University.}
\end{abstract}

\begin{document}
\maketitle

\section{Introduction}

Gun violence is a critical issue for many major cities.
In 2016, Chicago saw a 58\% surge in gun homicides and over 4,000 shooting victims, more than any other city comparable in size~\cite{kapustin2017gun}.
Recent data suggest that gun violence victims and perpetrators tend to have gang associations~\cite{kapustin2017gun}.
Notably, there were fewer homicides originating from physical altercations in 2016 than in the previous year, but we have little empirical evidence explaining why.
Burgeoning social science research indicates that gang violence may be exacerbated by escalation on social media and the ``digital street"~\cite{lane2016digital} where exposure to aggressive and threatening text and images
can lead to physical retaliation, a behavior known as ``Internet banging" or ``cyberbanging"~\cite{patton2013internet}.

Violence outreach workers present in these communities are thus attempting~\cite{rise2017} to prioritize their outreach around contextual features in social media posts indicative of offline violence, and to try to intervene and de-escalate the situation when such features are observed.
However, as most tweets do not explicitly contain features correlated with pathways of violence, an automatic or semi-automatic method that could flag a tweet as potentially relevant would lower the burden of this task.
The automatic interpretation of tweets or other social media posts could therefore be very helpful in intervention, but quite challenging to implement for a number of reasons, e.g. the informal language, the African American Vernacular English, and the potential importance of context to the meaning of the post. In specific communities (e.g. communities with high rates of violence) it can be hard even for human outsiders to understand what is actually going on.

\begin{figure}[t]
  \centering
  \includegraphics[width=\columnwidth]{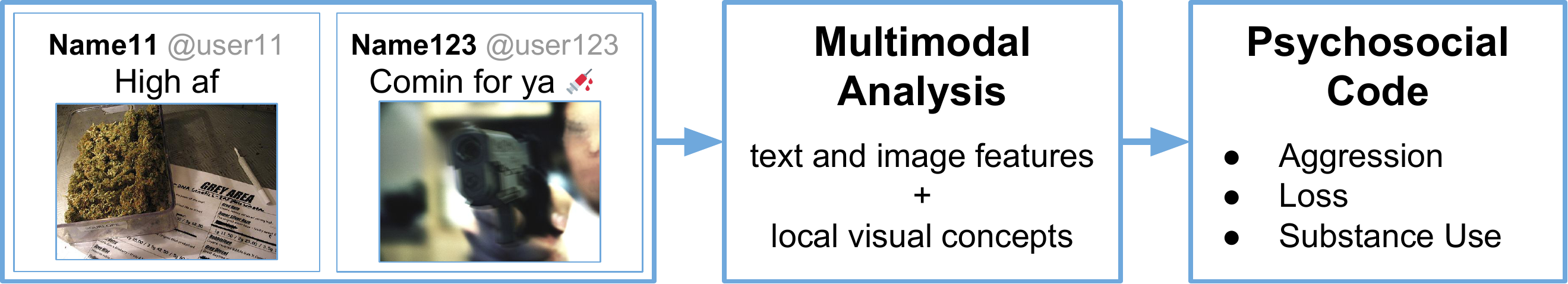}
\caption{We propose a multimodal system for detecting psychosocial codes of social media tweets\protect\footnotemark related to gang violence.}
  \label{fig:diagram}
\end{figure}
\footnotetext{Note that the ``tweets'' in Figure~\ref{fig:diagram} were created for illustrative purpose using Creative Commons images from Flickr and are NOT actual tweets from our corpus. Attributions of images in Figure~\ref{fig:diagram}, from left to right: ``IMG\_0032.JPG" by sashimikid, used under CC BY-NC-ND $2.0$, ``gun" by andrew\_xjy, used under CC BY-NC-ND $2.0$.}

To address this challenge, we have undertaken a first multimodal step towards developing such a system that we illustrate in Figure~\ref{fig:diagram}.
Our major contributions lie in innovative application of multimedia analysis of social media in practical social work study, specifically covering the following components:
\begin{itemize}
\item We have developed a rigorous framework to collect context-correlated tweets of gang-associated youth from Chicago containing images, and high-quality annotations for these tweets.
\item We have teamed up computer scientists and social work researchers to define a set of visual concepts of interest.
\item We have analyzed how the psychosocial codes \textit{loss}, \textit{aggression}, and \textit{substance use} are expressed in tweets with images and developed methods to automatically detect these codes, demonstrating a significant performance gain of 18\% by multimodal fusion.
\item We have trained and evaluated detectors for the concepts and psychosocial codes, and analyzed the usefulness of the local visual concepts, as well as the relevance of image vs. text for the prediction of each code.
\end{itemize}

\section{Related Work}

The City of Chicago is presently engaged in an attempt to use an algorithm to predict who is most likely to be involved in a shooting as either a victim or perpetrator \cite{ssl_chicago};  however, this strategy has been widely criticized due to lack of transparency regarding the algorithm \cite{schmidt2018,sheley2017} and the potential inclusion of  variables that may be influenced by racial biases present in the criminal justice system (e.g. prior convictions) \cite{bbc2017precrime,nellis2008reducing}.

In \cite{gerber2014115}, Gerber uses statistical topic modeling on tweets that have geolocation to predict how likely 20 different types of crimes are to happen in individual cells of a grid that covers the city of Chicago.
This work is a large scale approach for predicting future crime locations, while we detect codes in individual tweets related to future violence.
Another important difference is that \cite{gerber2014115} is meant to assist criminal justice decision makers,
whereas our efforts are community based and have solid grounding in social work research.

Within text classification, researchers have attempted to extract social events from web data including detecting police killings \cite{keith2017identifying}, incidents of gun violence \cite{pavlick2016gun}, and protests \cite{hanna2017mpeds}. However, these works primarily focus on extracting events from news articles and not on social media and have focused exclusively on the text, ignoring associated images.

The detection of local concepts in images has made tremendous progress in recent years, with recent detection methods~\cite{girshick2015fast,ren2017faster,rfcn2016,liu2016ssd,yolo2016} 
leveraging deep learning and efficient architecture enabling high quality and fast detections.
These detection models are usually trained and evaluated on datasets such as the PascalVOC~\cite{pascalVOC} dataset and more recently the MSCOCO~\cite{MSCOCO} dataset.
However, the classes defined in these datasets are for generic consumer applications and do not include the visual concepts specifically related to gang violence,
defined in section~\ref{sec:ontology}. 
We therefore need to define a lexicon of gang-violence related concepts
and train own detectors for our local concepts.

The most relevant prior work is that of \cite{blevins2016automatically}. They predict aggression and loss in the tweets of Gakirah Barnes and her top communicators using an extensive set of linguistic features, including mappings of African American vernacular English and emojis to entries in the Dictionary of Affective Language (DAL). The linguistic features are used in a linear SVM to make a 3-way classification between loss, aggression, and other. In this paper we additionally predict the presence of substance use, and model this problem as three binary classification problems since multiple codes may simultaneously apply. We also explore character and word level CNN classifiers, in addition to exploiting image features and their multimodal combinations.

\section{Dataset}

In this section we detail how we have gathered and annotated the data used in this work.

\subsection{Obtaining Tweets}

Working with community social workers, we identified a list of 200 unique users residing in Chicago neighborhoods with high rates of violence.
These users all suggest on Twitter that they have a connection, affiliation, or engagement with a local Chicago gang or crew. All of our users were chosen based on their connections to a seed user, Gakirah Barnes, and her top 14 communicators in her Twitter network\footnote{Top communicators were statistically calculated by most mentions and replies to Gakirah Barnes.}. Gakirah was a self-identified gang member in Chicago, before her death in April, 2014. Additional users were collected using snowball sampling techniques~\cite{enlighten37493}.
Using the public Twitter API, in February 2017 we scraped all obtainable tweets from this list of 200 users.
For each user we then removed all retweets, quote tweets and tweets without any image, limiting the number of remaining tweets per user to 20 to avoid most active users being overrepresented.
In total the resulting dataset consists of 1,851 tweets from 173 users.

\subsection{Local Visual Concepts\label{sec:ontology}} 

\begin{figure*}[t]
\centering
\subfloat[handgun, long gun]{\includegraphics[height=2.3cm]{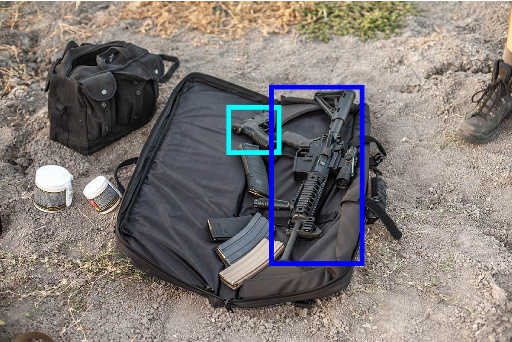}}
\hfill
\subfloat[person, hand gesture]{\includegraphics[height=2.3cm]{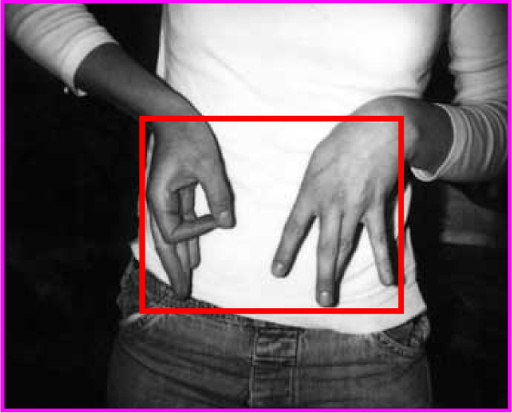}}
\hfill
\subfloat[money]{\includegraphics[height=2.3cm]{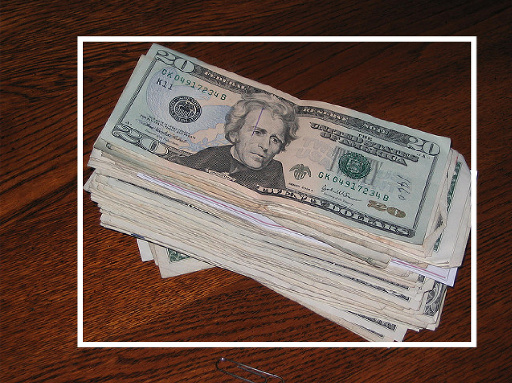}}
\hfill
\subfloat[marijuana, joint]{\includegraphics[height=2.3cm]{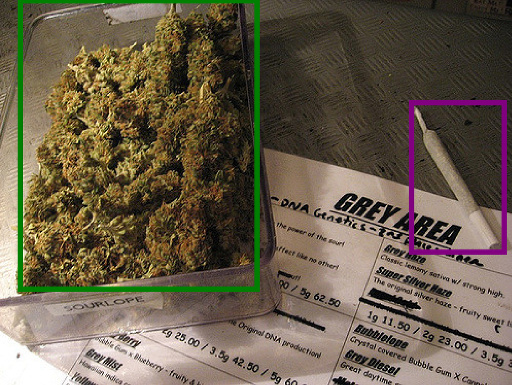}}
\hfill
\subfloat[lean]{\includegraphics[height=2.3cm]{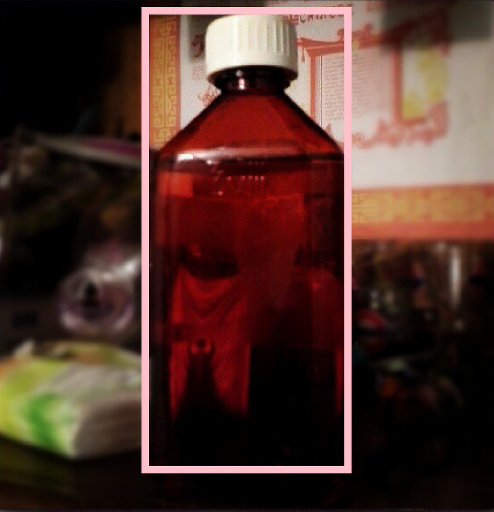}}
\hfill
\subfloat[person, tattoo]{\includegraphics[height=2.3cm]{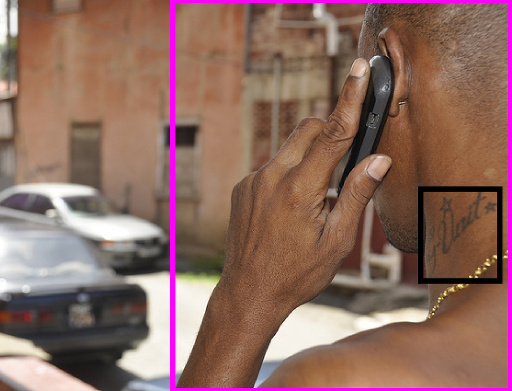}}
\caption{Examples of our gang-violence related visual concepts annotated on Creative Commons\protect\footnotemark images downloaded from Flickr.}
  \label{fig:ex_local_concepts}
\end{figure*}
\footnotetext{Attributions of Figure~\ref{fig:ex_local_concepts}, from left to right:
``GUNS" by djlindalovely, used under CC BY-NC-ND $2.0$, 
``my sistah the art gangstah" by barbietron, used under CC BY-NC $2.0$,
``Money" by jollyuk, used under CC BY $2.0$,
``IMG$\_$0032.JPG" by sashimikid, used under CC BY-NC-ND $2.0$,
``\#codeine time" by amayzun, used under CC BY-NC-ND $2.0$,
``G Unit neck tattoo, gangs Trinidad" by bbcworldservice, used under CC BY-NC $2.0$.
Each image has been modified to show the bounding boxes of the local concepts of interest present in it.}

To extract relevant information in tweet images related to gang violence, we develop a specific lexicon consisting of important and unique visual concepts often present in tweet images in this domain.
This concept list was defined through an iterative process involving discussions between computer scientists and social work researchers.
We first manually went through numerous tweets with images and discussed our observations to find which kind of information could be valuable to detect, either for direct detection of ``interesting" situations but also for extracting background information such as affiliation to a specific gang that can be visible from a tattoo.
Based on these observations we formulated a preliminary list of visual concepts.
We then collectively estimated utility (how useful is the extraction of the concept for gang violence prevention?), detectability (is the concept visible and discriminative enough for automatic detection?), and observability for reliable annotation (can we expect to obtain a sufficient number of annotations for the concept?), in order to refine this list of potential concepts and obtain the final lexicon.

Our interdisciplinary collaboration helped to minimize the risk of overseeing potentially important information
or misinterpreting behaviors that are specific to this particular community.
For example, on the images we frequently find people holding handguns with an extended clip and in many of these cases the guns are held at the clip only.
The computer scientists of our team did not pay much attention to the extended clips and were slightly confused by this way of holding the guns,
but then came to learn that in this community an extended clip counts as a sort of status symbol,
hence this way of holding is meant to showcase a common status symbol.
Such cross-disciplinary discussions lead to inclusion of concepts such as \textit{tattoos} and separation of concepts to \textit{handgun} and \textit{long gun} in our concept lexicon.

From these discussions we have derived the following set of local concepts (in image) of interest:
\begin{itemize}
\item General: \textit{person}, \textit{money}
\item Firearms: \textit{handgun}, \textit{long gun}
\item Drugs: \textit{lean}, \textit{joint}, \textit{marijuana}
\item Gang affiliation: \textit{hand gesture}, \textit{tattoo}
\end{itemize}

This list was designed in such a way that after the training process described above,  it could be further expanded (e.g. by specific hand gestures or actions with guns). We give examples of our local concepts in Figure~\ref{fig:ex_local_concepts}.

\subsection{Psychosocial Codes}

Prior studies~\cite{blevins2016automatically,patton2017gang} have identified \textit{aggression}, \textit{loss} and \textit{substance use} as emergent themes in initial qualitative analysis that were associated with Internet banging,
an emerging phenomenon of gang affiliates using social media to trade insults or make violence threats.
Aggression was defined as posts of communication that included an insult, threat, mentions of physical violence, or plans for retaliation. 
Loss was defined as a response to grief, trauma or a mention of sadness, death, or incarceration of a friend or loved one.
Substance use consists of mentions, and replies to images that discuss or show any substance (e.g. marijuana or a liquid substance colloquially referred to as ``lean", see example in Figure~\ref{fig:ex_local_concepts}) with the exception of cigarettes and alcohol. 

The main goal of this work is to automatically detect a tweet that can be associated with any or multiple of these three psychosocial codes (\textit{aggression}, \textit{loss} and \textit{substance use}) exploiting both textual and visual content.

\subsection{Annotation\label{sec:annotation}}

The commonly used annotation process based on crowd sourcing like Amazon Mechanical Turk is not suitable due to the special domain-specific context involved and the potentially serious privacy issues associated with the users and tweets.

Therefore, we adapted and modified the Digital Urban Violence Analysis Approach (DUVAA)~\cite{patton2016using,blevins2016automatically} for our project. 
DUVAA is a contextually-driven multi-step qualitative analysis and manual labeling process used for determining meaning in both text and images by interpreting both on- and offline contextual features. 
We adapted this process in two main ways. First, we include a step to uncover annotator bias through a baseline analysis of annotator perceptions of meaning. 
Second, the final labels by annotators undergo reconciliation and validation by domain experts living in Chicago neighborhoods with high rates of violence.
Annotation is provided by trained social work student annotators and domain experts, community members who live in neighborhoods from which the Twitter data derives. 
Social work students are rigorously trained in textual and discourse analysis methods using the adapted and modified DUVAA method described above. 
Our domain experts consist of Black and Latino men and women who affiliate with Chicago-based violence prevention programs. 
While our domain experts leverage their community expertise to annotate the Twitter data, our social work annotators undergo a five stage training process to prepare them for eliciting context and nuance from the corpus. 

We used the following tasks for annotation:
\begin{itemize}
\item In the \textit{bounding box annotation task}, annotators are shown the text and tweet of the image.
Annotators are asked to mark all local visual concepts of interest by drawing bounding boxes directly on the image.
For each image we collected two annotations.
\item To reconcile all conflicts between annotations we implemented a \textit{bounding box reconciliation task} where conflicting annotations are shown side by side and the better annotation can be chosen by the third annotator.
\item For \textit{code annotation}, tweets including the text, image and link to the original post, are displayed and for each of the three codes \textit{aggression}, \textit{loss} and \textit{substance use}, there is a checkbox the annotator is asked to check if the respective code applies to the tweet.
We collected two student annotations and two domain expert annotations for each tweet.
In addition, we created one extra code annotation to break ties for all tweets
with any disagreement between the student annotations.
\end{itemize}

Our social work colleagues took several measures to ensure the quality of the resulting dataset during the annotation process. Annotators met weekly as a group with an expert annotator to address any challenges and answer any questions that came up that week. This process also involved iterative correction of reoccurring annotation mistakes and infusion of new community insights provided by domain experts. Before the meeting each week, the expert annotator closely reviewed each annotator's interpretations and labels to check for inaccuracies.

During the annotation process, we monitored statistics of the annotated concepts.
This made us realize that for some visual concepts of interest, the number of expected instances in the final dataset was comparatively small.\footnote{We were aiming for at least around 100-200 instances for training plus additional instances for testing.}
Specifically, this affected the concepts \textit{handgun}, \textit{long gun}, \textit{money}, \textit{marijuana}, \textit{joint}, and \textit{lean}.
For all of these concepts we crawled additional images from Tumblr, using the public Tumblr API with a keyword-based approach for the initial crawling.
We then manually filtered the images we retrieved to obtain around 100 images for each of these specific concepts.
Finally we put these images into our annotation system and annotated them w.r.t. all local visual concepts listed in Section \ref{sec:ontology}.

\subsection{Statistics}

The distribution of concepts in our dataset is shown in Table \ref{tab:dataset_statistics}.
Note that in order to ensure sufficient quality of the annotations, but also due to the nature of the data, we relied on a special annotation process and kept the total size of the dataset comparatively small.

\begin{figure}
  \centering
  \includegraphics[width=0.9\columnwidth]{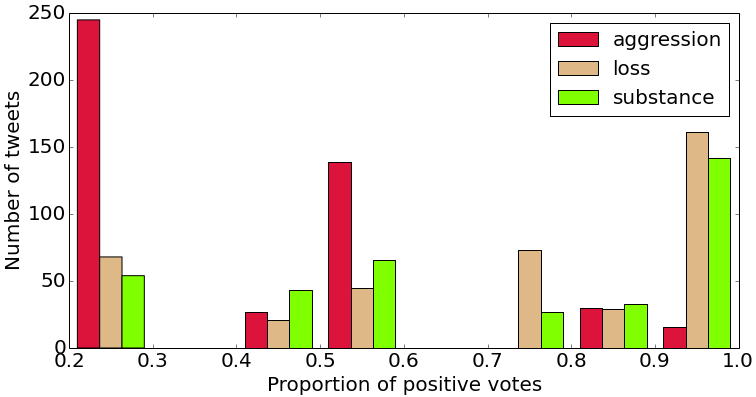}
  \caption{Annotator consensus for all psychosocial codes. For better visibility, we exclude tweets that were unanimously annotated as not belonging to the respective codes. Note that for each tweet there are 4 or 5 code annotations.}
  \label{fig:annotator_consensus}
\end{figure}

Figure~\ref{fig:annotator_consensus} displays the distributions of fractions of positive votes for all 3 psychosocial codes.
These statistics indicate that for the code \textit{aggression}, disagreement between annotators is substantially higher than for the codes \textit{loss} and \textit{substance use}, which both display a similar pattern of rather high annotator consensus.

\begin{table}
    \begin{tabular}{|c|c|c|c|}
        \hline
        {\bfseries Concepts/Codes} & {\bfseries Twitter} & {\bfseries Tumblr} & {\bfseries Total} \\
        \hline
        \textit{handgun} &
         164 & 41 & 205 \\
        \textit{long gun} &
         15 & 105 & 116\\
        \textit{joint} &
         185 & 113 & 298\\
        \textit{marijuana} &
         56 & 154 & 210\\
        \textit{person} &
         1368 & 74 & 1442\\
        \textit{tattoo} &
         227 & 33 & 260\\
        \textit{hand gesture} &
         572 & 2 & 574\\
        \textit{lean} &
         43 & 116 & 159\\
        \textit{money} &
         107 & 138 & 245\\
        \hline
        \textit{aggression} &
         457 (185)  & - & 457 (185)\\
        \textit{loss} &
         397 (308) & - & 397 (308)\\
        \textit{substance use} &
         365 (268) & - & 365 (268) \\
         \hline
    \end{tabular}
    \caption{Numbers of instances for the different visual concepts and psychosocial codes in our dataset.
    For  the different codes, the first number indicates for how many tweets at least one annotator assigned the corresponding code, numbers in parentheses are based on per-tweet majority votes.}
    \label{tab:dataset_statistics}
\end{table}

\subsection{Ethical considerations}

The users in our dataset comprise youth of color from marginalized communities in Chicago with high rates of gun violence. Releasing the data has the potential to further marginalize and harm the users who are already vulnerable to surveillance and criminalization by law enforcement. 
Thus, we will not be releasing the dataset used for this study.
However, to support research reproducibility, we will release only
the extracted linguistic and image features without revealing the
raw content; this enables other researchers to continue research
on training psychosocial code detection models without 
compromising the privacy of our users.
Our social work team members initially attempted to seek informed consent, but to no avail, as participants did not respond to requests.
To protect users, we altered text during any presentation so that tweets are not searchable on the Internet, excluded all users that were initially private or changed their status to private during the analysis, and consulted Chicago-based domain experts on annotation decisions, labels and dissemination of research.

\section{Methods\label{sec:methods} for Multimodal Analysis}

In this section we describe the building blocks for analysis, 
the text features and image features used as input for the psychosocial code classification with an SVM, 
and the multimodal fusion methods we explored.
Details of implementation and analysis of results will be presented in Sections~\ref{sec:experiments} and~\ref{sec:analysis}.

\subsection{Text features\label{sec:text_features}}

As text features, we exploit both sparse linguistic features as well as dense vector representations extracted from a CNN classifier operating at either the word or character level.

\subsubsection*{Linguistic features}
To obtain the linguistic features, we used the feature extraction code of \cite{blevins2016automatically} from which we obtained the following:
\begin{itemize}
\item Unigram and bigram features.
\item Part-of-Speech (POS) tagged unigram and bigram features. The POS tagger used to extract these features was adapted to this domain and cohort of users.
\item The minimum and maximum pleasantness, activation, and imagery scores of the words in the input text. These scores are computed by looking up each word's associated scores in the Dictionary of Affective Language (DAL).
Vernacular words and emojis were mapped to the Standard American English of the DAL using a translation phrasebook derived from this domain and cohort of users.
\end{itemize}

\subsubsection*{CNN features}

To extract the CNN features we train binary classifiers for each code.
We use the same architecture for both the word and character level models and so we describe only the word level model below. 
Our CNN architecture is roughly the same as \cite{kim2014convolutional} but with an extra fully connected layer before the final softmax. 
I.e., the text is represented as a sequence of embeddings, over which we run a series of varying width one-dimensional convolutions with max-pooling and a pointwise-nonlinearity; the resultant convolutional feature maps are concatenated and fed into a multi-layer perceptron (MLP) with one hidden layer and softmax output. After training the network, the softmax layer is discarded, and we take the hidden layer output in the MLP as the word or character feature vector to train the psychosocial code SVM.

\subsection{Image features\label{sec:image_features}}

We here describe how we extract visual features from the images that will be fed to the psychosocial code classifier.

\subsubsection*{Local visual concepts} 

To detect the local concepts defined in section~\ref{sec:ontology}, we adopt the Faster R-CNN model~\cite{ren2017faster}, a state-of-the-art method for object detection in images. 
The Faster R-CNN model introduced a \emph{Region Proposal Network} (RPN) to produce region bounds and objectness score at each location of a regular grid. The bounding boxes proposed by the RPN are fed to a Fast R-CNN~\cite{girshick2015fast} detection network.
The two networks share their convolutional features, enabling the whole Faster R-CNN model to be trained end-to-end and to produce fast yet accurate detections.
Faster R-CNN has been shown~\cite{huang2017speed} to be one of the best models among the modern convolutional object detectors in terms of accuracy.
Details on the training of the model on our data are provided in Section~\ref{sec:exp-local_concepts}.
We explore the usefulness of the local visual concepts in two ways:
\begin{itemize}
\item For each \textit{local visual concept} detected by the faster R-CNN, we count the frequency of the concept detected in a given image. For this, we only consider predictions of the local concept detector with a confidence higher than a given threshold, which is varied in experiments.
\item In order to get a better idea of the potential usefulness of our proposed local visual concepts,  we add one model to the experiments that uses \textit{ground truth local concepts} as features. This corresponds to features from a perfect local visual concept detector. This method is considered out-of-competition and is not used for any fusion methods. It is used only to gain a deeper understanding of the relationship between the local visual concepts and the psychosocial codes.
\end{itemize}

\subsubsection*{Global features}

As \textit{global image features} we process the given images using a deep convolutional model (Inception-v3 \cite{inceptionv3}) pre-trained on ImageNet \cite{imagenet} and use activations of the last layer before the classification layer as features.
We decided not to update any weights of the network due to the limited size of our dataset and because such generic features have been shown to have a strong discriminative power~\cite{razavian2014cnn}.

\subsection{Fusion methods for code detection}

In addition to the text- and image-only models that can be obtained by using individually each feature described in Sections~\ref{sec:text_features} and~\ref{sec:image_features}, we evaluate several tweet classification models that combine multiple kinds of features from either one or both modalities.
These approaches always use features of all non-fusion methods for the respective modalities outlined in Sections~\ref{sec:text_features} and~\ref{sec:image_features}, and combine information in one of the following two ways:
\begin{itemize}
\item  \textit{Early fusion}: the different kinds of features are concatenated into a single feature vector, which is then fed into the SVM.
For example, the text-only early fusion model first extracts linguistic features and deploys a character and a word level CNN to compute two 100-dimensional representations of the text, and then feeds the concatenation of these three vectors into an SVM for classification.
\item \textit{Late fusion} corresponds to an ensemble
approach. Here, we first train separate SVMs on the code classification task for each feature as input, and then train another final SVM to detect the psychosocial codes from the probability outputs of the previous SVMs.
\end{itemize}

\section{Experiments\label{sec:experiments}}

Dividing by twitter users\footnote{We chose to do the split on a user basis so that tweets of the same user are not repeated in both training and test sets.}, we randomly split our dataset into 5 parts with similar code distributions and total numbers of tweets. We use these splits for 5-fold cross validation, 
i.e. all feature representations that can be trained and the psychosocial code prediction models are trained on 4 folds and tested on the unseen 5th fold.
All reported performances and sensitivities are averaged across these 5 data splits.
Statements on statistical significance are based on 95\% confidence intervals computed from the 5 values on the 5 splits.

We first detail how the text and image representations are trained on our data.
We then discuss the performance of different uni- and multimodal psychosocial code classifiers.
The last two experiments are designed to provide additional insights into the nature of the code classification task and the usefulness of specific concepts.

\subsection{Learning text representations\label{sec:exp_learn_text}}

\subsubsection*{Linguistic features}

We do not use all the linguistic features described in Section~\ref{sec:text_features} as input for the SVM but instead
during training apply feature selection using an ANOVA F-test that selects the top $1,300$ most important features. 
Only the selected features are provided to the SVM for classification.
We used the default SVM hyperparameter settings of~\cite{blevins2016automatically}.
\subsubsection*{CNN features}
We initialize the word embeddings with pretrained 300-dimensional \textit{word2vec} \cite{word2vec} embeddings.\footnote{\url{https://code.google.com/p/word2vec/}} For the character level model, we used 100-dimensional character embeddings randomly initialized by sampling uniformly from $(-0.25,0.25)$.
In both CNN models we used convolutional filter windows of size 1 to 5 with 100 feature maps each. The convolutional filters applied in this way can be thought of as word (or character) ngram feature detectors, making our models sensitve to chunks of one to five words (or characters) long. We use a 100-dimensional hidden layer in the MLP. 
During cross-validation we train the CNNs using the Nesterov Adam \cite{dozat2016incorporating} optimizer with a learning rate of .002, early stopping on 10\% of the training fold, and dropout of .5 applied to the embeddings and convolutional feature maps.

\subsection{Learning to detect local concepts\label{sec:exp-local_concepts}}

Our local concepts detector is trained using the image data from Twitter and Tumblr and the corresponding bounding box annotations. We use the Twitter data splits defined above and similarly define five splits for the Tumblr data with similar distribution of concepts across different parts. 
We train a Faster R-CNN\footnote{We use the publicly available implementation from: \url{https://github.com/endernewton/tf-faster-rcnn}} model using a 5-fold cross validation, training using 4 splits of the Twitter and Tumblr data joined as a training set.
We evaluate our local concepts detection model on the joined test set, as well as separately on the Twitter and Tumblr test set, and will discuss its performance in section~\ref{sec:analysis_local_concepts}.

The detector follows the network architecture of VGG-16 and is trained using the 4-step alternating training approach detailed in~\cite{ren2017faster}. The network is initialized with an ImageNet-pretrained model and trained for the task of local concepts detection. We use an initial learning rate of $0.001$ which is reduced by a factor of $0.9$ every 30k iterations and trained the model for a total of 250k iterations. We use a momentum of $0.8$ and a weight decay of $0.001$.

During training, we augment the data by flipping images horizontally. In order to deal with class imbalance while training, we weigh the classification cross entropy loss for each class by the logarithm of the inverse of its proportion in the training data. We will discuss in detail the performance of our detector in Section~\ref{sec:analysis_local_concepts}.

\begin{table*}[!th]
\begin{adjustbox}{max width=\textwidth}
\setlength\tabcolsep{4pt}
    \begin{tabular}{|c|c|c|| c|c|c|c |c|c|c|c| c|c|c|c|| c|}
        \hline
        \multirow{2}{*}{\bfseries Modality} & \multirow{2}{*}{\bfseries Features} & \multirow{2}{*}{\bfseries Fusion} 
& \multicolumn{4}{c|}{\bfseries Aggression} & \multicolumn{4}{c|}{\bfseries Loss} & \multicolumn{4}{c||}{\bfseries Substance use} & 
\multirow{2}{*}{\bfseries mAP}
         \\
        \cline{4-15}
         & & & P & R & F1 & AP & P & R & F1 & AP & P & R & F1 & AP & 
         \\
        \hline
        - & - (random baseline) & - &
        	0.25   &   0.26  &  0.26 & 0.26&
        	0.17    &  0.17 & 0.17 & 0.20 & 
            0.18    &  0.18 & 0.18 & 0.20 & 0.23 \\
        - & - (positive baseline) & - &
        	0.25  &    1.00 &   0.40 & 0.25 &
            0.21  &    1.00 & 0.35&  0.22 &
            0.20   &   1.00&  0.33 & 0.20 & 0.22 \\
        \hline
        text & linguistic features & - & 
        	0.35   &   0.34 & 0.34 & 0.31 &  
            0.71   &   0.47&  0.56 & 0.51 & 
            0.25   &   0.53 & 0.34 & 0.24 & 0.35 \\
        text & CNN-char & - & 
        	0.37 &     0.47 &   0.39 & 0.36& 
            0.75 &     0.66 & 0.70 & \alert{\textbf{0.77}} & 
            0.27    &  0.32 & 0.29&  0.28 & 0.45 \\
        text & CNN-word & - & 
        	0.39  &    0.46  &  0.42 & 0.41& 
            0.71   &   0.65&  0.68& \alert{\textbf{0.77}} & 
            0.28    &  0.30 & 0.29 & 0.31 & 0.50 \\
            \hline
        text & all textual & early & 
        	0.40  &    0.46 & 0.43&  0.42& 
            0.70  &    0.73&  0.71 & \alert{\textbf{0.81}} & 
            0.25   &   0.37&  0.30&  0.30 & 0.51 \\
        text & all textual & late & 
        	0.43   &   0.41 & 0.42&  0.42& 
            0.69   &   0.65 & 0.67 & \alert{\textbf{0.79}} & 
            0.29   &   0.37 & 0.32 & 0.32 & 0.51 \\
        \hline
        image & inception global & - & 
        	0.43  &    0.64 &  0.51 & \alert{\textbf{0.49}} &
            0.38   &   0.57&  0.45 & 0.43 &
            0.41   &   0.62 & 0.49 & \alert{\textbf{0.48}} & 0.47 \\
        image & Faster R-CNN local (0.1) & - & 
        	0.43 &     0.64 &   0.52&  \alert{\textbf{0.47}} & 
            0.28  &    0.56 & 0.37 & 0.31& 
            0.44   &   0.30&  0.35 & 0.37 & 0.38 \\
        image & Faster R-CNN local (0.5) & - & 
        	0.47  &    0.48  &  0.47 & 0.44&  
            0.30  &    0.39&  0.33 & 0.31& 
            0.46   &   0.12&  0.19 & 0.30  & 0.35 \\
            \hline
        image & all visual & early & 
        	0.49   &   0.62 & 0.55 &  \alert{\textbf{0.55}}*& 
            0.38  &    0.57 & 0.45 & 0.44& 
            0.41   &   0.59 & 0.48 & \alert{\textbf{0.48}} & 0.49 \\
        image & all visual & late & 
        	0.48  &    0.51 & 0.49 &  \alert{\textbf{0.52}} & 
            0.40  &    0.51 & 0.44 &  0.43 & 
            0.47  &    0.52 & 0.50 & \alert{\textbf{0.51}}* & 0.49 \\
        \hline
        image+text & all textual + visual & early &
        	0.48   &   0.51 & 0.49 & \alert{\textbf{0.53}}&  
            0.72  &    0.73&  0.73 &  \alert{\textbf{0.82}}* &
            0.37   &   0.53 & 0.43 & \alert{\textbf{0.45}} & \alert{\textbf{0.60}} \\
        image+text & all textual + visual & late &
        	0.48  &    0.44&  0.46 & \alert{\textbf{0.53}} &  
            0.71   &   0.67&  0.69 & \alert{\textbf{0.80}} &
            0.44   &   0.43&  0.43& \alert{\textbf{0.48}} & \alert{\textbf{0.60}}* \\
        \hline
    \end{tabular}
\end{adjustbox}
\caption{Results for detecting the psychosocial codes: aggression, loss and substance use.
    For each code we report precision (P), recall (R), F1-scores (F1) and average precision (AP). Numbers shown are mean values of 5-fold cross validation performances. 
    The highest performance (based on AP) for each code is marked with an asterisk. 
In bold and red we highlight all performances not significantly worse than the highest one (based on statistical testing with 95\% confidence intervals).}
    \label{tab:results_code}
\end{table*}

\subsection{Detecting psychosocial codes}

We detect the three psychosocial codes separately, i.e. for each code we consider the binary classification task of deciding whether the code applies to a given tweet.

For our experiments we consider a tweet to belong to the positive class of a certain code if at least one annotator marked the tweet as displaying that code.
For the negative class we used all tweets 
that were not marked by any annotator as belonging to the code
(but might belong or not belong to any of the two other codes).
We chose this way of converting multiple annotations to single binary labels
because our final system is not meant to be used as a fully automatic detector
but as a pre-filtering mechanism for tweets that are potentially useful for social workers.
Given that the task of rating tweets with respect to such psychosocial codes inevitably depends on the perspective on the annotator to a certain extent, we think that even in case of a majority voting mechanism, important tweets might be missed.\footnote{For future work we are planning to have a closer look at the differences between annotations of community experts and students and based on that treat these types of annotations differently. We report a preliminary analysis in that direction in Section~\ref{sec:analysis_annotation}.}

In addition to the models trained using the features described in Section \ref{sec:methods}, we also evaluate two baselines that do not process the actual tweet data in any way.
Our \textit{random baseline} uses the training data to calculate the prior probability of a sample belonging to the positive class and for each test sample predicts the positive class with this probability without using any information about the sample itself.
The other baseline, \textit{positive baseline}, always outputs the positive class.

All features except the linguistic features were fed to an SVM using the RBF kernel for classifying the psychosocial codes.
For linguistic features, due to issues when training with an RBF kernel, we used a linear SVM 
with squared hinge loss, as in \cite{blevins2016automatically},
and C = $0.01$, $0.03$ and $0.003$ for detecting \textit{aggression}, \textit{loss} and \textit{substance use} respectively.
Class weight was set to balanced, with all other parameters kept at their default values.
We used the SVM implementation of the Python library scikit-learn \cite{scikit-learn}.
This two stage approach of feature extraction plus classifier was chosen to allow for a better understanding of the contributions of each feature.
We preferred SVMs in the 2nd stage over deep learning methods since SVMs can be trained on comparatively smaller datasets without the need to optimize many hyperparameters.

For all models we report results with respect to the following metrics:
precision, recall and F1-score (always on positive class), and average precision (using detector scores to rank output).
The former 3 measures are useful to form an intuitive understanding of the performances,
but for drawing all major conclusions we rely on average precision, 
which is an approximation of the area under the entire precision-recall curve, as compared to measurement at only one point.

The results of our experiments are shown in Table \ref{tab:results_code}.
Our results indicate that image and text features play different roles in detecting different psychosocial codes. Textual information clearly dominates the detection of code \textit{loss}.
We hypothesize that loss is better conveyed textually whereas substance use and aggression are easier to express visually. Qualitatively, the linguistic features with the highest magnitude weights (averaged over all training splits) in a linear SVM bear this out, with the top five features for loss being i) \textit{free}, ii) \textit{miss}, iii) \textit{bro}, iv) \textit{love} v) \textit{you}; the top five features for substance use being i) \textit{smoke}, ii) \textit{cup}, iii) \textit{drank}, iv) \textit{@mention} v) \textit{purple}; and the top five features for aggression being i) \textit{Middle Finger Emoji}, ii) \textit{Syringe Emoji}, iii) \textit{opps}, iv) \textit{pipe} v) \textit{2017}. 
The loss features are obviously related to the death or incarceration of a loved one (e.g. \textit{miss} and \textit{free} are often used in phrases wishing someone was freed from prison). 
The top features for aggression and substance use are either emojis which are themselves pictographic representations, i.e. not a purely textual expression of the code, or words that reference physical objects (e.g. \textit{pipe, smoke, cup}) which are relatively easy to picture.

Image information dominates classification of both the \textit{aggression} and \textit{substance use} codes.
Global image features tend to outperform local concept features, but combining local concept features with global image features achieves the best image-based code classification performance. 
Importantly, by fusing both image and text features, the combined detector performs consistently very well for all three codes, with the mAP over three codes being $0.60$, compared to $0.51$ for the text only detector and $0.49$ for the image only detector.
This demonstrates a relative gain in mAP of around 20\% of the multimodal approach over any single modality.

\subsection{Sensitivity analysis\label{sec:exp_sensitivity_analysis}}

We performed additional experiments to get a better understanding of the usefulness of our local visual concepts for the code prediction task.
For sensitivity analysis we trained linear SVMs on psychosocial code classification, using as features either the local visual concepts detected by Faster R-CNN or the ground truth visual concepts.
All reported sensitivity scores are average values of the corresponding coefficients of the linear SVM, computed across the 5 folds used for the code detection experiments.
Results from this experiment can be found in Table~\ref{tab:sensitivity_analysis}.
\begin{table}
\begin{adjustbox}{max width=\columnwidth}
\setlength\tabcolsep{3pt}
    \begin{tabular}{|c| c|c|c|c|c|c|c|c|c|}
        \hline
        \multirow{2}{*}{\bfseries Concept} & \multicolumn{3}{c|}{\bfseries Aggression}
        & \multicolumn{3}{c|}{\bfseries Loss}
        & \multicolumn{3}{c|}{\bfseries Substance use} \\
        \cline{2-10}
          & 0.1 & 0.5 & GT & 0.1 & 0.5 & GT & 0.1 & 0.5 & GT  \\
        \hline
        \textit{handgun} &
         0.73 & 0.93 & 1.05 &
          0.06 & 0.10 & 0.06 &
         0.06 & 0.09 & 0.11  \\
        \textit{long gun} &
         0.26 & 0.91 & 1.30 &
          -0.17 & 0.14 & 0.14 &
         0.42 & 0.04 & -0.47 \\
        \textit{joint} &
         0.42 & -0.08 & 0.05 &
          -0.15 & 0.00 & 0.10 &
         0.25 & 1.3 & 1.41 \\
        \textit{marijuana} &
         0.17 & 0.18 & 0.12 &
          -0.19 & -0.45 & -0.35 &
         0.93 & 1.29 & 1.47  \\
        \textit{person} &
         0.34 & -0.01 & -0.17 &
          0.11 & 0.10 & 0.12 &
         0.04 & 0.28 & -0.01 \\
        \textit{tattoo} &
         -0.11 & -0.09 & 0.01 &
          -0.02 & 0.03 & -0.03 &
         0.04 & 0.06 & -0.02 \\
        \textit{hand gesture} &
         0.20 & 0.67 & 0.53 &
          -0.01 & 0.12 & 0.05 &
         0.01 & 0.06 & -0.02 \\
        \textit{lean} &
         -0.07 & 0.03 & -0.28 &
          -0.20 & -0.06 & -0.14 &
         0.68 & 0.59 & 1.46 \\
        \textit{money} &
         -0.06 & 0.06 & -0.02 &
          0.00 & -0.01 & -0.01 &
         0.18 & -0.04 & -0.19 \\
        \hline
        \hline
        F1 &
         0.51  & 0.46 &0.65   &
          0.37 & 0.33 & 0.38 &
         0.34 & 0.17 & 0.76  \\
         AP &
         0.41 & 0.39 & 0.54 &
          0.29 & 0.28 & 0.30 &
         0.33 & 0.27 & 0.72   \\
         \hline
    \end{tabular}
    \end{adjustbox}
    \caption{Sensitivity of visual local concept based classifiers w.r.t. the different concepts. For each of the three psychosocial codes, we include two versions that use detected local concepts (``0.1" and ``0.5", where the number indicates the detection score threshold) and one version that uses local concept annotations as input (``GT").}
    \label{tab:sensitivity_analysis}
\end{table}

From classification using ground truth visual features we see that
for detecting \textit{aggression}, the local visual concepts \textit{handgun} and \textit{long gun} are important,
while for detecting \textit{substance use}, the concepts \textit{marijuana}, \textit{lean}, \textit{joint} are most significant.
For the code \textit{loss}, \textit{marijuana} as the most relevant visual concept correlates negatively with \textit{loss}, but overall, significance scores are much lower.

Interestingly, the model that uses the higher detection score threshold of $0.5$ for the local visual concept
detection behaves similarly to the model using ground truth annotations, even though the classification performance is better with the lower threshold.
This could indicate that using a lower threshold makes the code classifier learn to exploit false alarms of the concept detector.

However, it needs to be mentioned that sensitivity analysis can only measure how much the respective classifier uses the different parts of the input, given the respective overall setting.
This can give you useful information about which parts are \textit{sufficient} for obtaining comparable detection results, but there is no guarantee that the respective parts are
also \textit{necessary} for achieving the same classification
performance.
\footnote{For example, imagine that two hypothetical concepts A and B correlate perfectly with a given class and a detector for this class is given both concepts as input. The detector could make its decision based on A alone, but A is not really necessary since the same could be achieved by using B instead.}

For this reason, we ran an ablation study to get quantitative measurements on the necessity of local visual concepts for code classification.

\subsection{Ablation study\label{sec:exp_ablation_study}}

In our ablation study we repeated the psychosocial code classification experiment using ground truth local visual concepts as features, excluding one concept at a time to check how this affects overall performance of the model.

We found that for \textit{aggression}, removing the concepts \textit{handgun} or \textit{hand gesture} leads to the biggest drops in performance, while for \textit{substance use}, the concepts \textit{joint}, \textit{marijuana} and \textit{lean} are most important.
For \textit{loss}, removal of none of the concepts causes any significant change.
See Table~\ref{tab:ablation_results} for further details.
\begin{table}[t]
    \begin{tabular}{|c| c|c|c|c|}
        \hline
        \multirow{2}{*}{\bfseries Removed concept} & \multicolumn{2}{c|}{\bfseries Aggression} & \multicolumn{2}{c|}{\bfseries Substance use} \\
        \cline{2-5}
          & F1 & AP & F1 & AP  \\
        \hline
        \textit{handgun} &
        \textbf{-0.10}  & \textbf{-0.15} &
        \textbf{-0.01}  &  0.01 \\
        \textit{long gun} &
        -0.01    & \textbf{-0.01} &
        -0.00 & -0.00 \\
        \textit{joint} &
        0.00 &  -0.00 &
        \textbf{-0.35}  & \textbf{-0.28} \\
        \textit{marijuana} &
         0.00    &  0.00 &
        \textbf{-0.09}  & \textbf{-0.09} \\
        \textit{person} &
        \textbf{-0.01}   & -0.01 &
        -0.01  & -0.00 \\
        \textit{tattoo} &
        0.00    & 0.00 &
        0.01   & -0.00 \\
        \textit{hand gesture} &
        \textbf{-0.13}    & \textbf{-0.09}&
        0.00   &  0.00 \\
        \textit{lean} &
        -0.00  &  0.00 &
        \textbf{-0.07}    & \textbf{-0.07} \\
        \textit{money} &
        0.00  &  0.00 &
        0.00   &  0.00 \\
        \hline
    \end{tabular}
    \caption{Differences in psychosocial code detection performance of detectors with specific local concepts removed as compared to a detector that uses all local concept annotations.
    (Numbers less than 0 indicate that removing the concept reduces the corresponding score.) Bold font indicates that the respective number is significantly less than 0. For the code loss none of the numbers was significantly different from 0, hence we decided to not list them in this table.}
    \label{tab:ablation_results}
\end{table}

\section{Open challenges\label{sec:analysis}}

In this section, we provide a more in-depth analysis of what makes our problem especially challenging and how we plan to address those challenges in the future.

\subsection{Local concepts analysis\label{sec:analysis_local_concepts}}

We report in Table~\ref{tab:local_concepts_det} the average precision results of our local concept detection approach on the ``Complete" test set, i.e. joining data from both Twitter and Tumblr, and separately on the Twitter and Tumblr test sets. 
We compute the average precision on each test fold separately and report the average and standard deviation values over the 5 folds.
When looking at the results on the ``Complete" test set, we see average precision values ranging from $0.26$ on \textit{tattoo} to $0.80$ for \textit{person} and the mean average precision of $0.54$ indicating a rather good performance. 
This results on the ``Complete" test set hides two different stories, however, as the performance is much lower on the Twitter test set (mAP of $0.29$) than on the Tumblr one (mAP of $0.81$).

\begin{table}[t]
    \begin{tabular}{|c|c|c|c|}
        \hline
        \multirow{2}{*}{\bfseries Concept} & {\bfseries Complete} & {\bfseries Twitter} & {\bfseries Tumblr} \\
        \cline{2-4}
          & AP $\pm$ SD & AP $\pm$ SD & AP $\pm$ SD \\
 \hline
 \textit{handgun} & 0.30 $\pm$ 0.07 & 0.13 $\pm$ 0.02 & 0.74 $\pm$ 0.11 \\
 \textit{long gun} & 0.78 $\pm$ 0.03 & 0.29 $\pm$ 0.41 & 0.85 $\pm$ 0.05 \\
 \textit{joint} & 0.30 $\pm$ 0.07 & 0.01 $\pm$ 0.01 & 0.57 $\pm$ 0.04 \\
 \textit{marijuana} & 0.73 $\pm$ 0.08 & 0.28 $\pm$ 0.17 & 0.87 $\pm$ 0.09 \\
 \textit{person} & 0.80 $\pm$ 0.03 & 0.80 $\pm$ 0.03 & 0.95 $\pm$ 0.03 \\
 \textit{tattoo} & 0.26 $\pm$ 0.06 & 0.08 $\pm$ 0.02 & 0.84 $\pm$ 0.06 \\
 \textit{hand gesture} & 0.27 $\pm$ 0.05 & 0.28 $\pm$ 0.04 & 0.83 $\pm$ 0.29 \\  \textit{lean} & 0.78 $\pm$ 0.07 & 0.38 $\pm$ 0.15 & 0.87 $\pm$ 0.03 \\
 \textit{money} & 0.60 $\pm$ 0.02 & 0.35 $\pm$ 0.08 & 0.73 $\pm$ 0.05 \\
 \hline
 \hline
 mAP & 0.54 $\pm$ 0.01 & 0.29 $\pm$ 0.05 & 0.81 $\pm$ 0.02 \\
 \hline
    \end{tabular}
    \caption{Local concepts detection performance.}
    \label{tab:local_concepts_det}
\end{table}

As detailed in Section~\ref{sec:annotation}, we have crawled additional images, especially targeting the concepts with a low occurrence count in Twitter data as detailed in Table~\ref{tab:dataset_statistics}. 
However, crawling images from Tumblr targeting keywords related to those concepts lead us to gather images where the target concept is the main subject in the image, while in our Twitter images they appear in the image but are rarely the main element in the picture. 
Further manually analyzing the images crawled from Twitter and Tumblr, we have confirmed this ``domain gap" between the two sources of data that can explain the difference of performance.
This puts in light the challenges associated with detecting these concepts in our Twitter data. 
We believe the only solution is therefore to gather additional images from Twitter from similar users. 
This will be part of the future work of this research.

The local concepts are highly relevant for the detection of the codes \textit{aggression} and \textit{substance use} as it can be highlighted in the column GT in Table~\ref{tab:sensitivity_analysis} and from the ablation study reported in Table~\ref{tab:ablation_results}.
The aforementioned analysis of the local concepts detection limitation on the Twitter data explains why the performance using the detected concepts
is substantially lower than when using ground truth local concepts.
We will therefore continue to work on local concepts detection in the future as we see they could provide significant help in detecting these two codes and also because they would help in providing a clear interpretability of our model.

\subsection{Annotation analysis\label{sec:analysis_annotation}}
In order to identify factors that led to divergent classification between social work annotators and domain experts, we reviewed 10\% of disagreed-upon tweets with domain experts.
In general, knowledge of local people, places, and behaviors accounted for the majority of disagreements. In particular, recognizing and having knowledge of someone in the image (including their reputation, gang affiliation, and whether or not they had been killed or incarcerated) was the most common reason for disagreement between our annotators and domain experts. 
Less commonly, identifying or recognizing physical items or locations related to the specific cultural context of the Chicago area (e.g., a home known to be used in the sale of drugs) also contributed to disagreement. The domain experts' nuanced understanding of hand signs also led to a more refined understanding of the images, which variably increased or decreased the perceived level of aggression. 
For example, knowledge that a certain hand sign is used to disrespect a specific gang often resulted in increased perceived level of aggression. 
In contrast, certain hand gestures considered to be disrespectful by our social work student annotators (e.g., displaying upturned middle fingers) were perceived to be neutral by domain experts and therefore not aggressive.
Therefore, continuous exchange with the domain experts is needed to always ensure that the computer scientists are aware of all these aspects when further developing their methods.

\subsection{Ethical implications}
Our team was approached by violence outreach workers in Chicago to begin to create a computational system that would enhance violence prevention and intervention.
Accordingly, our automatic vision and textual detection tools were created
to assist social workers in their efforts to understand and prevent community violence through social media, but
not to optimize any systems of surveillance.
This shift away from identifying potentially violent users to understanding pathways to violent online content highlights systemic gaps in economic, educational, and health-related resources that are often root causes to violent behavior. 
Our efforts for ethical and just treatment of the users who provide our data include encryption of all Twitter data, removal of identifying information during
presentation of work (e.g., altering text to eliminate searchability), and the inclusion of Chicago-based community members as domain experts in the analysis and validation of our findings. 
Our long term efforts include using multimodal analysis to enhance current violence prevention efforts by providing insight into social media behaviors that may shape future physical altercations.

\section{Conclusion}

We have introduced the problem of multimodal social media analysis for gang violence prevention
and presented a number of automatic detection experiments
to gain insights into the expression of \textit{aggression}, \textit{loss} and \textit{substance use} in tweets coming from this specific community,
measure the performance of state-of-the-art methods on detecting these codes in tweets that include images,
and analyze the role of the two modalities text and image in this multimodal tweet classification setting.

We proposed a list of general-purpose local visual concepts and
showed that despite insufficient performance of current local concept detection,
when combined with global visual features,
these concepts can help visual detection of \textit{aggression} and \textit{substance use} in tweets.
In this context we also analyzed in-depth the contribution of all individual concepts.

In general, we found the relevance of the text and image modalities in tweet classification to depend heavily on the specific code being detected,
and demonstrated that combining both modalities leads to a significant improvement of overall performance across all 3 psychosocial codes.

Findings from our experiments affirm prior social science research indicating that youth use social media to respond to, cope with, and discuss their exposure to violence. Human annotation, however, remains an important element in vision detection in order to understand the culture, context and nuance embedded in each image.
Hence, despite promising detection results,
we argue that psychosocial code classification is far from being solved by automatic methods.
Here our interdisciplinary approach clearly helped to become aware of the whole complexity of the task, but also to see the broader context of our work, including important ethical implications which were discussed above.

\section*{Acknowledgments}
During his stay at CU, the first author was supported by a fellowship within the FITweltweit programme of the German Academic Exchange Service (DAAD).
Furthermore, we thank all our annotators: Allison Aguilar, Rebecca Carlson, Natalie Hession, Chloe Martin, Mirinda Morency.

\bibliographystyle{ACM-Reference-Format}
\bibliography{bibliography}

\end{document}